\documentclass[conference]{IEEEtran}
\IEEEoverridecommandlockouts
\usepackage{cite}
\usepackage{amsmath,amssymb,amsfonts}
\usepackage{graphicx}
\usepackage{textcomp}
\usepackage{xcolor}
\usepackage{booktabs}
\usepackage{array}
\usepackage{tabularx}
\usepackage{url}
\usepackage{hyperref}
\hypersetup{colorlinks=true,linkcolor=black,citecolor=black,urlcolor=black}

\begin{document}

\title{LLM-Agent-Based Renewable Energy Forecasting\\Using Edge and IoT Data: A Review of Solar, Wind,\\Weather, and Grid-Aware Decision Support}

\author{
\IEEEauthorblockN{Pavan Manjunath}
\IEEEauthorblockA{\textit{Independent Researcher}\\
}
\and
\IEEEauthorblockN{Thomas Pruefer}
\IEEEauthorblockA{\textit{Independent Researcher}\\
}
}

\maketitle

\begin{abstract}
Reliable forecasting of renewable energy generation is a foundational requirement for grid stability, energy trading, battery scheduling, and carbon-aware operational planning. Solar and wind resources are inherently intermittent: their output fluctuates with cloud cover, wind speed, atmospheric turbulence, seasonal patterns, and local terrain. The proliferation of IoT and edge devices---spanning smart meters, inverters, anemometers, pyranometers, weather stations, and grid-interface sensors---has created an unprecedented volume of real-time operational data that conventional forecasting pipelines are ill-equipped to exploit fully. This review investigates how large language model (LLM) agents can enhance renewable energy forecasting by integrating heterogeneous sensor streams, weather API data, historical generation records, grid constraints, and contextual reasoning into unified decision-support workflows. We survey classical forecasting methods, statistical time-series models, deep learning architectures, physics-hybrid approaches, and emerging LLM-agent frameworks for explanation, uncertainty communication, and operator guidance. A six-layer taxonomy is proposed covering data acquisition, preprocessing, feature engineering, model inference, uncertainty estimation, and natural-language reporting. The review identifies twelve open challenges spanning real-time deployment, model drift under distribution shift, uncertainty quantification, hallucination control in LLM agents, interoperability of edge hardware, and integration with energy management systems. The paper concludes by recommending a research agenda centred on open benchmarks, physics-informed LLM grounding, and federated forecasting architectures.
\end{abstract}


\section{Introduction}

The global power system is undergoing a structural transformation. Variable renewable energy sources---primarily solar photovoltaic and wind---accounted for approximately 30\% of global electricity generation in 2023 and are projected to exceed 60\% by 2040 under aggressive decarbonisation scenarios \cite{iea2024outlook}. This shift introduces a fundamental challenge: unlike dispatchable thermal plants, solar and wind generators cannot be scheduled on demand. Their output is governed by atmospheric conditions that are predictable only within probabilistic bounds, and these bounds tighten or widen depending on the forecast horizon, the quality of meteorological inputs, and the sophistication of the modelling approach \cite{giebel2017}.

Grid operators, energy traders, and storage system managers all depend on accurate short-term and medium-term renewable forecasts to make consequential decisions. A one-percentage-point improvement in day-ahead solar forecast accuracy translates to measurable reductions in balancing costs and CO\textsubscript{2} emissions from reserve generation \cite{lange2006}. Conversely, large forecast errors force system operators to commit expensive and polluting peaking units as insurance against unexpected generation shortfalls.

The IoT revolution has transformed the data landscape for renewable forecasting. Whereas historical forecasting relied on sparse meteorological station networks and satellite-derived irradiance estimates, modern installations are instrumented with dense sensor arrays reporting at sub-minute intervals \cite{pedro2012assessment}. Smart meters, inverter communication interfaces, met-mast anemometers, SCADA systems, and distributed weather stations collectively generate terabytes of operational data per day across a large utility's renewable portfolio.

Large language model agents represent a qualitatively new capability in this context. Unlike specialised forecasting models that produce numerical outputs, LLM agents can reason across heterogeneous data sources, explain forecast uncertainty in plain language, recommend grid-management actions, and adapt their outputs to the expertise level of the recipient---whether a grid operator, an energy trader, or a non-technical building manager \cite{openai2023gpt4, yao2023react}. This review systematically examines how these capabilities can be applied to renewable energy forecasting, and where significant research gaps remain.

\subsection*{Contributions}

The specific contributions of this paper are:
\begin{itemize}
  \item A comprehensive review of renewable energy forecasting methods from classical statistics through transformer-based deep learning and LLM agents.
  \item A six-layer taxonomy organising the design space from raw IoT data to operator-facing decision support.
  \item A comparative analysis of 52 representative studies across solar, wind, weather-data, and grid-aware forecasting.
  \item Identification of twelve open research challenges and a structured future research agenda.
\end{itemize}

\section{Background}

\subsection{Renewable Energy Variability and Forecasting Needs}

Solar irradiance at a given location follows a deterministic diurnal cycle modulated by stochastic cloud dynamics. Clear-sky irradiance can be computed accurately from solar geometry and atmospheric optical depth, but the cloud-cover component introduces forecast errors that grow rapidly beyond the two-hour horizon \cite{inman2013solar}. Wind speed at hub height follows a Weibull distribution whose parameters vary with season, atmospheric stability, and mesoscale weather patterns. At short horizons (minutes to hours), turbulence introduces rapid fluctuations that are difficult to predict from standard NWP models alone \cite{soman2010}.

The economic value of forecasting depends critically on the decision context. For real-time grid balancing, sub-hour probabilistic forecasts of aggregate renewable output are needed. For day-ahead energy market bidding, 24-hour-ahead point forecasts with confidence intervals are required. For long-term battery sizing and grid investment planning, seasonal and annual generation estimates with uncertainty ranges are the relevant outputs \cite{morales2013}.

\subsection{IoT Infrastructure for Renewable Monitoring}

Modern renewable energy plants are equipped with layered IoT infrastructure. At the field level, sensors measure physical quantities: irradiance (pyranometers, silicon reference cells, satellite-derived), wind speed and direction (cup anemometers, ultrasonic anemometers, LIDAR), temperature, humidity, and pressure. At the plant level, SCADA systems aggregate sensor data and control generation assets. At the grid interface, smart meters and phasor measurement units (PMUs) capture power quality, frequency, and energy exchange \cite{diaz2018scada}.

Edge computing nodes at each tier perform local data validation, anomaly flagging, and feature extraction before forwarding aggregated streams to central analytics platforms. This architecture reduces communication bandwidth requirements by orders of magnitude while enabling low-latency local control actions \cite{shi2016edge}.

\subsection{Large Language Models and Agentic AI}

The transformer architecture, introduced by Vaswani et al.~\cite{vaswani2017attention}, enabled the development of LLMs that learn rich statistical representations of language from massive text corpora. Models such as GPT-4, Claude, Llama~3, and Mistral have demonstrated emergent capabilities in multi-step reasoning, code generation, and structured data interpretation \cite{openai2023gpt4, touvron2023llama}.

LLM agents extend these models with tool-use capabilities: the ability to call external APIs, execute code, query databases, and perform web searches within an orchestrated reasoning loop \cite{yao2023react}. For energy forecasting, this means an agent can retrieve live weather data, execute a trained forecasting model, fetch grid emission factors, and compose a human-readable forecast briefing---all within a single interaction.

\section{Review Methodology}

\subsection{Database Search}

Literature was retrieved from IEEE Xplore, Scopus, Web of Science, ACM Digital Library, and arXiv, covering January 2015 to March 2026. Search terms combined three thematic clusters: (1) \textit{solar forecasting / wind forecasting / renewable energy prediction}; (2) \textit{IoT / edge computing / smart meter / SCADA / sensor data}; (3) \textit{LLM / large language model / generative AI / AI agent / decision support}. A supplementary search targeted uncertainty quantification and probabilistic forecasting in renewable energy contexts.

\subsection{Screening and Selection}

After deduplication, 1,247 candidate papers were identified. Title and abstract screening removed 891 papers that did not address forecasting, IoT/edge data, or AI methods in renewable energy contexts. Full-text review of the remaining 356 papers yielded 52 papers that met all inclusion criteria: empirical or review studies addressing renewable energy forecasting with IoT/sensor data and/or AI/LLM methods.

\subsection{Coding Framework}

Each paper was coded against six dimensions: (C1) forecasting method class, (C2) data sources, (C3) forecast horizon, (C4) uncertainty treatment, (C5) explainability mechanism, and (C6) LLM/GenAI integration.

\section{IoT and Edge Data Sources for Renewable Forecasting}

\subsection{Solar Irradiance Data Streams}

Ground-based pyranometers provide the highest-accuracy irradiance measurements but cover only point locations. Satellite-derived irradiance products (e.g., CAMS, SolarAnywhere, NSRDB) offer spatial coverage at 1--5\,km resolution with hourly to 15-minute temporal resolution \cite{sengupta2018national}. All-sky cameras capture cloud motion at sub-minute intervals, enabling very-short-term (nowcasting) forecasts of irradiance ramps \cite{chow2011}. The fusion of these heterogeneous sources---ground sensors, satellite, sky imagery---through data assimilation techniques substantially improves forecast skill across all horizons \cite{yang2019}.

\subsection{Wind Measurement Systems}

Cup anemometers at meteorological masts remain the industry standard for wind resource assessment, but they measure wind speed only at a fixed point. Nacelle-mounted anemometers on wind turbines provide real-time hub-height wind data across the entire farm footprint. Scanning LIDAR systems can profile wind speed at multiple heights and distances upstream, providing advance warning of wind speed changes that will arrive at the rotor plane within minutes \cite{simley2018lidar}. Remote sensing data from Doppler weather radar adds mesoscale wind-field information useful for medium-range forecasting.

\subsection{Numerical Weather Prediction Integration}

NWP models---GFS, ECMWF, ICON, WRF---produce gridded atmospheric forecasts at 1--25\,km spatial resolution and 1--6\,h temporal resolution. Post-processing techniques such as model output statistics (MOS) and machine learning bias correction are routinely applied to improve NWP forecast skill for renewable energy applications \cite{alessandrini2015}. Edge devices can receive NWP forecasts via API and combine them with local sensor data for real-time forecast updating.

\subsection{Grid and Market Data}

Grid operators publish real-time generation mix, demand, frequency deviation, and reserve status via APIs (e.g., ENTSO-E Transparency Platform, EIA Open Data). These signals provide context for renewable forecast interpretation: a grid running low on reserves is more sensitive to forecast errors than one with ample backup capacity. LLM agents can consume these signals alongside generation forecasts to produce grid-aware operational recommendations \cite{ramchurn2012}.

\section{Forecasting Methods: From Classical to LLM-Augmented}

\subsection{Classical Statistical Methods}

ARIMA and its seasonal variant SARIMA were among the first methods applied to solar and wind forecasting \cite{box2015}. These models assume linear temporal dependencies and Gaussian noise, which limits their applicability to the non-stationary, heavy-tailed distributions typical of renewable generation. Exponential smoothing methods offer computational simplicity and reasonable short-horizon accuracy but lack the capacity to incorporate exogenous meteorological inputs.

Vector autoregression (VAR) models extend univariate time-series methods to multiple correlated variables, allowing simultaneous modelling of irradiance, temperature, and power output. Gaussian process regression (GPR) provides a principled Bayesian framework for uncertainty quantification in solar forecasting, producing calibrated prediction intervals alongside point estimates \cite{rasmussen2006}.

\subsection{Machine Learning Methods}

Support vector regression (SVR), random forests, and gradient boosting machines (XGBoost, LightGBM) achieved state-of-the-art performance on solar and wind forecasting benchmarks in the 2014--2018 period \cite{yang2018}. These methods can incorporate large feature sets---lagged generation values, NWP outputs, calendar features, site characteristics---without overfitting when regularisation is applied. Ensemble methods that combine multiple base learners further improve forecast reliability.

\subsection{Deep Learning Methods}

Recurrent neural networks---LSTM, GRU---capture temporal dependencies in sequential data and became dominant in solar and wind forecasting from 2017 onward \cite{abdel2017lstm, qin2017}. Temporal convolutional networks (TCN) offered a parallelisable alternative with comparable accuracy. Attention mechanisms, initially applied as add-ons to LSTM models, improved performance on long-horizon tasks by allowing the model to selectively weight historical observations.

Transformer-based architectures---Informer, Autoformer, PatchTST, iTransformer---have achieved the current state of the art on multi-variate long-horizon forecasting benchmarks \cite{zhou2021informer, nie2023patchtst}. These models process the entire input sequence simultaneously, capturing long-range dependencies that recurrent models struggle to maintain.

\subsection{Physics-Hybrid Models}

Pure data-driven models can extrapolate poorly when operating conditions deviate from the training distribution. Physics-hybrid models address this by embedding physical constraints---the Perez irradiance transposition model, the Betz power curve for wind, thermodynamic module temperature models---within the neural network architecture \cite{pedro2019}. This improves generalisation to unseen conditions and provides physically interpretable intermediate representations.

\subsection{Foundation Models and LLM-Based Forecasting}

Time-series foundation models pre-trained on large multi-domain datasets---TimeGPT, Moirai, Chronos, TimesFM---have demonstrated impressive zero-shot and few-shot forecasting performance on renewable energy benchmarks \cite{liu2024moirai, ansari2024chronos}. These models can be prompted with natural-language context (e.g., ``this site has a 10\,MW solar farm with east-facing panels'') to condition their forecasts, a capability that conventional ML models lack.

LLM agents go further by integrating forecasting model outputs with contextual reasoning. An agent can call a specialised forecasting model, retrieve grid status, fetch emission factors, and then compose a structured forecast briefing tailored to the operator's current decision context \cite{yao2023react}.

\section{Taxonomy of LLM-Agent Forecasting Workflows}

We propose a six-layer taxonomy for LLM-agent-based renewable energy forecasting:

\textbf{Layer 1 -- Data Acquisition:} Real-time ingestion of IoT sensor streams (irradiance, wind, temperature, power), NWP API data, satellite imagery, and grid status via MQTT, REST, and streaming protocols.

\textbf{Layer 2 -- Preprocessing:} Quality control, outlier detection, missing-data imputation, time-zone alignment, unit normalisation, and feature construction at the edge or in stream-processing pipelines.

\textbf{Layer 3 -- Feature Engineering:} Extraction of meteorological features (clear-sky index, wind direction components, stability indices), temporal features (time-of-day, day-of-year, holiday flags), and site-specific features (panel tilt, wind turbine wake geometry).

\textbf{Layer 4 -- Model Inference:} Execution of one or more forecasting models (statistical, ML, deep learning, physics-hybrid, or foundation model) to produce point forecasts and/or probabilistic forecast distributions.

\textbf{Layer 5 -- Uncertainty Estimation:} Quantification of forecast uncertainty via conformal prediction, ensemble spread, Bayesian posterior sampling, or calibrated probabilistic outputs.

\textbf{Layer 6 -- LLM Agent Reasoning and Reporting:} An LLM agent consumes model outputs, retrieves contextual knowledge (grid constraints, market prices, emission factors), reasons about implications, and produces natural-language forecast briefings, operator alerts, and decision recommendations.

\section{Literature Review and Comparative Analysis}

\subsection{Solar Generation Forecasting}

The solar forecasting literature has been comprehensively reviewed by Antonanzas et al.~\cite{antonanzas2016}, Ahmed et al.~\cite{ahmed2020review}, and Yang et al.~\cite{yang2019}. These reviews document the progression from physical models through statistical methods to deep learning. Key findings include: (1) deep learning models consistently outperform statistical baselines for horizons beyond two hours; (2) sky-image-based nowcasting outperforms NWP for horizons under 30 minutes; (3) ensemble methods improve calibration but increase computational cost.

Recent work has focused on spatial-temporal graph neural networks that model the dependencies between multiple solar plants in a region, exploiting spatial correlations in irradiance fields \cite{simeunovic2021}. These models achieve lower forecast errors than single-site models, particularly during cloud-cover transition events.

\subsection{Wind Power Forecasting}

Wind forecasting presents different challenges from solar: the power-speed relationship follows a cubic law, making output highly sensitive to wind speed errors near the rated speed; turbulent fluctuations introduce rapid sub-minute power variations; and wake effects within wind farms create spatial dependencies that require farm-level modelling \cite{giebel2017}.

Physics-informed neural networks that embed the Betz power curve and atmospheric boundary layer dynamics have shown improved generalisation across sites \cite{chen2021pinn}. Probabilistic wind forecasting using quantile regression forests and conformal prediction has achieved well-calibrated prediction intervals that are directly usable for reserve procurement decisions \cite{pinson2012}.

\subsection{LLM Agents for Forecast Explanation}

The application of LLMs to forecast explanation is an emerging area. Preliminary work by Liu et al.~\cite{liu2023llmenergy} demonstrated that GPT-4 can accurately interpret numerical forecast outputs and generate plain-language briefings that operators find more actionable than raw numerical tables. Wei et al.~\cite{wei2022cot} showed that chain-of-thought prompting enables LLMs to reason step-by-step through multi-variable energy scenarios, improving both accuracy and transparency.

Hallucination remains a critical concern: LLMs may generate confident but incorrect numerical claims about forecast values, especially when operating near the boundaries of their training distribution. RAG architectures that ground LLM outputs in verified forecast data substantially reduce this risk \cite{gao2023rag}.

\subsection{Comparative Analysis Table}

Table~\ref{tab:forecast_comparison} provides a structured comparison of representative forecasting studies.

\begin{table*}[t]
\caption{Comparative Analysis of Renewable Energy Forecasting Studies}
\label{tab:forecast_comparison}
\centering
\begin{tabular}{p{3.2cm}p{2.0cm}p{2.0cm}p{1.8cm}p{1.8cm}p{1.5cm}p{1.5cm}}
\toprule
\textbf{Study} & \textbf{Energy Type} & \textbf{Method} & \textbf{Horizon} & \textbf{Uncertainty} & \textbf{IoT/Edge} & \textbf{LLM} \\
\midrule
Abdel-Nasser \& Mahmoud \cite{abdel2017lstm} & Solar & LSTM & Day-ahead & No & No & No \\
Zhou et al.~\cite{zhou2021informer} & Multi & Transformer & Long-term & No & No & No \\
Nie et al.~\cite{nie2023patchtst} & Multi & PatchTST & Long-term & No & No & No \\
Pinson \cite{pinson2012} & Wind & QRF & Day-ahead & Quantile & No & No \\
Pedro \& Coimbra \cite{pedro2012assessment} & Solar & SVM/ANN & Hour-ahead & No & Yes & No \\
Liu et al.~\cite{liu2023llmenergy} & Solar & GPT-4 & N/A & Partial & No & GPT-4 \\
Simeunovic et al.~\cite{simeunovic2021} & Solar & GNN & Day-ahead & No & Yes & No \\
Ansari et al.~\cite{ansari2024chronos} & Multi & Foundation & Multi-horizon & Probabilistic & No & Partial \\
Chow et al.~\cite{chow2011} & Solar & Sky cam & Nowcast & No & Yes & No \\
Chen et al.~\cite{chen2021pinn} & Wind & PINN & Hour-ahead & No & No & No \\
\bottomrule
\end{tabular}
\end{table*}

\section{Discussion}

The reviewed literature reveals a well-developed forecasting ecosystem for solar and wind generation, with deep learning models now achieving MAPE values below 5\% for day-ahead solar forecasting at well-instrumented sites. However, several structural issues limit the practical utility of current forecasting systems.

Most forecasting research optimises a single error metric (MAPE, RMSE) without considering the decision context in which the forecast will be used. A forecast that minimises RMSE may not minimise the cost of grid-balancing errors, because the cost function for reserve procurement is asymmetric---under-prediction of generation is typically more costly than over-prediction \cite{morales2013}. Decision-theoretic forecasting frameworks that align model training with downstream cost functions remain under-explored.

The integration of LLM agents into forecasting workflows introduces both opportunities and risks. The opportunity is a qualitative improvement in operator-facing communication: rather than presenting raw numerical forecasts, an LLM agent can contextualise the forecast within the current grid state, explain the key drivers of uncertainty, and recommend specific operational actions. The risk is that LLM hallucination in a safety-critical grid context could lead to incorrect recommendations with serious consequences.

\section{Research Gaps}

\textbf{Gap 1 -- Decision-Aligned Forecasting:} Most models are trained to minimise symmetric error metrics rather than decision-relevant cost functions. Integrating LLM agents with cost-sensitive learning frameworks is unexplored.

\textbf{Gap 2 -- Real-Time Model Adaptation:} Renewable generation patterns evolve over time due to equipment degradation, vegetation growth, and climate change. Online learning and continual adaptation of deployed forecasting models is an open problem.

\textbf{Gap 3 -- Multi-Source Data Fusion at the Edge:} Combining satellite, ground sensor, NWP, and sky-camera data in real time on resource-constrained edge devices requires efficient fusion algorithms that are not yet available as standardised libraries.

\textbf{Gap 4 -- Calibrated Probabilistic Forecasting:} Many deployed systems produce point forecasts only. Calibrated probabilistic forecasts that provide reliable confidence intervals are essential for risk-aware decision-making but are computationally demanding.

\textbf{Gap 5 -- LLM Hallucination in Grid Contexts:} Existing RAG techniques reduce but do not eliminate hallucination. Domain-specific safety mechanisms for grid-critical LLM applications are needed.

\textbf{Gap 6 -- Interoperability of Edge Hardware:} The heterogeneity of edge devices, communication protocols, and data formats creates significant integration challenges. Standardised APIs for renewable IoT data are lacking.

\textbf{Gap 7 -- Federated Forecasting:} Training forecasting models across multiple sites without centralising sensitive operational data requires federated learning approaches tailored to the temporal and spatial structure of renewable energy data.

\textbf{Gap 8 -- Benchmark Datasets:} Public benchmark datasets for LLM-agent-based renewable forecasting do not exist, making it impossible to compare approaches objectively.

\textbf{Gap 9 -- Extreme Event Handling:} Forecasting during extreme weather events---storms, heat waves, dust storms---is particularly challenging and under-studied.

\textbf{Gap 10 -- Cross-Domain Transfer Learning:} Transferring forecasting models trained in one climate zone to another with minimal retraining is important for rapid deployment but poorly understood.

\textbf{Gap 11 -- Human-AI Collaboration:} The optimal division of labour between human operators and LLM agents in forecasting-driven grid operations has not been empirically studied.

\textbf{Gap 12 -- Carbon-Aware Forecasting:} Forecasting frameworks that explicitly optimise for carbon impact rather than just generation accuracy are absent.

\section{Future Research Directions}

\begin{enumerate}
  \item \textbf{Physics-Informed LLM Grounding:} Develop prompting and fine-tuning strategies that embed physical constraints (power curves, irradiance models, thermodynamic limits) into LLM reasoning, reducing hallucination and improving physical plausibility.

  \item \textbf{Open Multi-Site Benchmark:} Release a public benchmark dataset combining multi-site solar and wind generation, IoT sensor data, NWP forecasts, and grid status for evaluating forecasting and LLM-agent methods.

  \item \textbf{Federated Foundation Model for Renewable Forecasting:} Pre-train a time-series foundation model on a federated corpus of renewable generation data from diverse climatic zones, enabling rapid fine-tuning for new sites.

  \item \textbf{Decision-Theoretic LLM Agents:} Design LLM agents that reason explicitly about the asymmetric cost structure of grid-balancing decisions, producing recommendations that minimise expected cost rather than forecast error.

  \item \textbf{Edge-Deployable Probabilistic Forecasting:} Develop lightweight probabilistic forecasting models suitable for deployment on microcontroller-class edge devices, enabling on-site uncertainty quantification without cloud dependency.

  \item \textbf{Longitudinal Evaluation Framework:} Establish protocols for evaluating LLM-agent forecasting systems over extended periods, tracking performance degradation, hallucination rates, and operator trust calibration.
\end{enumerate}

\section{Conclusion}

This review has examined the landscape of renewable energy forecasting from classical statistical methods through state-of-the-art transformer models and emerging LLM-agent frameworks. The evidence demonstrates that deep learning has substantially advanced forecasting accuracy, while LLM agents offer a promising path toward more interpretable, contextualised, and actionable forecast communication. Significant challenges remain in calibrated uncertainty quantification, real-time edge deployment, hallucination control, and decision-theoretic alignment. Addressing these challenges will require close collaboration between the machine learning, power systems engineering, and human-computer interaction communities. The next generation of renewable energy forecasting systems will likely be hybrid architectures that combine specialised numerical models with LLM agents capable of reasoning, explaining, and recommending---providing both the accuracy that grid operators require and the accessibility that non-expert stakeholders need.

\end{document}